# Sentiment analysis in Tourism: Fine-tuning BERT or sentence embeddings concatenation?

Ibrahim BOUABDALLAOUI
LASTIMI Laboratory, EST Salé
Mohammed V University in Rabat
Salé, Morocco
bd.ibrahim@hotmail.com

Fatima GUEROUATE
LASTIMI Laboratory, EST Salé
Mohammed V University in Rabat
Salé, Morocco
guerouate@gmail.com

Samya BOUHADDOUR
LASTIMI Laboratory, EST Salé
Mohammed V University in Rabat
Salé, Morocco
samya.bouhaddour@uit.ac.ma

Chaimae SAADI
LASTIMI Laboratory, EST Salé
Mohammed V University in Rabat
Salé, Morocco
chaimaesaadi900@gmail.com

Mohammed SBIHI
LASTIMI Laboratory, EST Salé
Mohammed V University in Rabat
Salé, Morocco
mohamed.sbihi@yahoo.fr

*Abstract*— Undoubtedly that the Bidirectional Encoder representations from Transformers is the most powerful technique in making Natural Language Processing tasks such as Named Entity Recognition, Question & Answers or Sentiment Analysis, however, the use of traditional techniques remains a major potential for the improvement of recent models, in particular word tokenization techniques and embeddings, but also the improvement of neural network architectures which are now the core of each architecture. recent. In this paper, we conduct a comparative study between Fine-Tuning the Bidirectional Encoder Representations from Transformers and a method of concatenating two embeddings to boost the performance of a stacked Bidirectional Long Short-Term Memory-Bidirectional Gated Recurrent Units model; these two approaches are applied in the context of sentiment analysis of shopping places in Morocco. A search for the best learning rate was made at the level of the two approaches, and a comparison of the best optimizers was made for each sentence embedding combination with regard to the second approach.

## I. Introduction

Sentiment analysis involves the evaluation of emotions, attitudes and opinions. Organizations and brands utilize this strategy to acquire important experiences into a client's reaction to a specific product or service. Sentiment analysis tools use advanced artificial intelligence technologies such as natural language processing, machine learning techniques, text analytics and data science to identify, extract and study subjective information. Fundamentally, opinion mining is utilized to group text as sure, negative or impartial. Traditional performance measurements, such as views, shares, snaps, likes or comments center around numbers. Sentiment analysis goes beyond quantity, as it focuses on the quality of the interaction between the audience and the subject. Sentiment analysis is particularly important in the analysis of tourist flows on social networks. This is where potential tourists immediately react to posts about a tourist site and even interact with each other, which is usually done in a sincere and direct way. If the feedback indicates a negative or erroneous perception of products or services, decision makers can quickly adjust and re-evaluate specific aspects. Technically, it is quite easy to analyze large amounts of text. However, the key is to distinguish between relevant and irrelevant text. For relevant reviews on the tourism site itself, the content should be filtered and divided. For example, depending on whether the text is really about a review of a tourist market or whether the review is about an entire city. Although this type of review is also very useful, it does not really contribute to the analysis if it is evaluated together with the tourist shopping place itself.

## II. Related work

An Automated Concatenation of Embeddings (ACE) had been proposed [1] to mechanize the most common way of finding better links of embeddings for organized expectation undertakings, in view of a definition enlivened by recent advancement on neural architecture search, specifically a controller on the other hand tests a connection of embeddings, as per its flow conviction of the viability of individual implanting types in thought for a task, and updates the conviction in light of a reward. The undertaking model predicts the task yield, while the controller looks for better embedding concatenation as the word portrayal for the task model to accomplish higher performances. For the competitors of embeddings on English datasets, the language explicit model for ELMo [2], Flair[3], base BERT [4], GloVe word embeddings [5], fastText word embeddings [6], noncontextual character embeddings [7], multilingual Flair (M-Flair), M-BERT [8] and XLM-R embeddings [9] had been utilized for this proposed

Research reported is supported by Al-Khawarizmi program of the CNRST and ADD to promote artificial intelligence projects in Moroccan industries.

approach, and had been tested and compared with two baselines in one hand: the first baseline is to let the task model learn by itself the contribution of each embedding candidate that is helpful to the task, the second baseline, a random search had been launched, with a strong baseline in Neural Architecture Searching (NAS) [10], and thus, It's been ran with the same maximum iteration as in ACE. In another hand, a BiLSTM with MaxPooling had been introduced [11] to show that adding an ordered progression of BiLSTM and max pooling layers yields cutting edge results for the SNLI sentence encoding-based models and the SciTail dataset, as well as provides strong results for the MultiNLI dataset. The neural architecture follows a sentence implanting based approach for Natural Language Inference (NLI) [12], the model contains sentence embeddings for the two info sentences, where the result of the sentence embeddings are consolidated using a heuristic [13], assembling the link (u, v), outright component wise contrast |u − v|, and component wise item u ∗ v. The joined vector is then given to a 3-layered multi-facet perceptron (MLP) with a 3-way SoftMax classifier.

III. METHODOLOGY AND RESULTS

*A. Dataset*

The dataset in this study is collected from TripAdvisor website, due to the huge amount of reviews in the tourist context, and also because of its dependability and improved content. Using a web scraping technique, we prevailed at gathering all reviews from the shopping section in TripAdvisor, including rates and location of every Moroccan shopping place recorded in the website. Shopping places known in TripAdvisor are 91, collected with their names and the normal of rates and number of reviews. Reviews are assembled in each shopping place name, the reviews vary between 1 review to more than 20000, depending the popularity of the place. Each shopping place has a quick presentation, pictures, address and nearby places, reviews and "Question & Answers" section. The shape of the collected data is 26244 rows and 4 major columns: "Name of the shop place", "Title of the review", "Review", "Rate". The name of shopping place may be duplicated in function of reviews number. The plot below shows the top 5 of the most frequent shopping places that has much reviews:

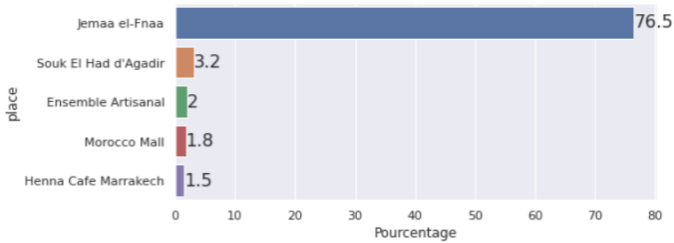

Fig. 1. Top-5 reviews counting following shopping place name

As can be seen "Jemaa El-Fena" shopping place took the first rank with 76,5% of the total of reviews collected from Shopping places section in TripAdvisor, this feature is considered as an outlier due to the unbalanced number of reviews in comparison with the other shopping places. Thus, in this study, we decided to focus only on the samples that have "Jemaa El-Fena" as a shopping place name, to set the constraint of various shopping places, and discover strengths and weaknesses of this touristic place via the feelings of the reviewers on the website. For the rate column, a conversion work is done from stars notation to numbers, and then to string labels: "bad" for 0, "neutral" for 1, "good" for 2.

*B. Fine-tuning BERT*

One of the revolutions in Natural Language Processing (NLP), is the appearance of transformers in which are considered an important reason why the machine learning community considers the BERT model (Bidirectional Encoder Representations from Transformers) as an essential technical innovation, hence the modeling linguistics of this model promises to take machine learning to new levels. One of the main reasons for the good performance of BERT on different NLP tasks was its utilization in semi-supervised learning. This means that the model is trained for a specific task that allows it to understand language patterns. After training, the BERT model has language processing capabilities that can be used to reinforce other models that we build and train using supervised learning. The usage of BERT in sentiment analysis has proven high performance [14] in many contexts such as business or politics…, which could encourage researchers to test the model in a tourist context to take profit from the richness of the data in tourism forums and show as a result, hidden insights which can be deducted from the prediction of sentiments towards a particular feature in a shopping place. One of the core strengths of BERT is the embedding part that it is used to extract features [15], and make sentence embedding vectors, from text. These embeddings are valuable for keyword or search purposes, semantic search and data retrieval. For instance, to match tourist's inquiries or searches against already answered questions or well documented searches, these portrayals will help him accurately recover results matching the traveler's goal and logical significance, regardless of whether there's no keyword or expression cross-over. In this regard, our approach is to build a based uncased BERT model with a maximum sequence length of 512 and define the multiclass labels that should be predicted (i.e. bad, neutral and good), and then split the data to 90% for training set and 10% for testing. A work of searching the optimal learning rate had been done by setting the optimizer to Adam and the loss function to categorical cross entropy, and so, a loop was incorporated on the learning rate parameter, ranging from $10^{-7}$ to $10^{-1}$, and the criterion to choose the best learning rate is to reach the lowest loss score, learning is achieved on 3 epochs, results can be seen in the plot below:

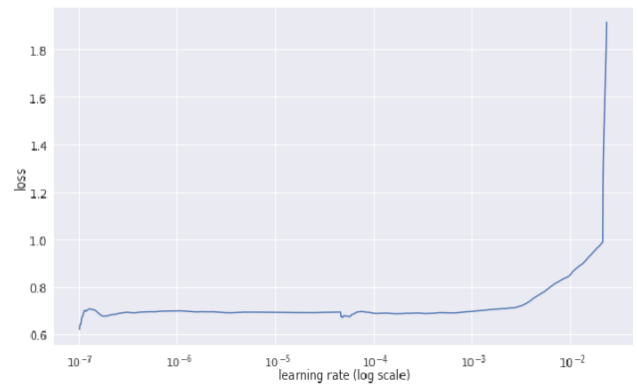

Fig. 2. Learning rate optimization curve

These results show that the best learning rate is the one that reached $2*10^{-4}$ with a loss value of 0.70, which is a preliminary score [16] that can be developed if we launch a BERT model while setting the optimal learning rate and increasing the number of epochs.

*C. Sentence embedding concatenation*

The text embedding was constructed using the average of the words contained in the short text of the enhanced title. The word embeddings are obtained by using obtained by using the whole corpus for training. Sentence embedding is a technique that is used many years in Natural Language Processing in order to perform an encoding textual architecture that have to be useful for mathematical processing (e.g. neural networks, Markovian systems, …). However, the use of these embedding sentences may be insufficient if we have a large corpus that contains several topics [17], and consequently, a text encoding problem will be invoked, and it is possible that the learning on this corpus may be biased [18]. For this, an alternative can be applied to overcome this problem: it is to develop a module of loading of embeddings that has as parameters two different embeddings, a dictionary of words, a dictionary of lemmatized words and the size of the vector of embeddings, subsequently we browse each defined embeddings and we store them in dictionaries that it will be stacked and we calculate the mean values, Then, we create an embedding matrix that will be refined according to the words that are in the corpus and the words that are defined in the parameters (dictionary of words and lemmatization dictionary that had been tokenized from the collected text using the "en_core_web_lg" Spacy model while defining indexes for each token). From a detailed spot, we can split the concatenation algorithm in 2 parts: the first part consists in preparing the data structures that will allow us to accommodate the two embedding layers to be concatenated, first, we have to calculate the size of the corpus, and define a vector that will contain the embeddings of the unknown words, this vector has the size of the embedding defined in the parameters and is initialized by zeros, then we load the first embedding that must be browsed using a dictionary, and store it in a NumPy array, and then calculate the mean of this array and store the results; similarly for the second embedding, then we initialize the embedding matrix with a Zero Matrix of a row size that is equal to the corpus length, and a column size that is equal to the embedding size. For the second part, it aims at filling the embedding matrix with the elements of the corpus by using the word dictionary (which is the elements of the corpus and its indexes), the defined embeddings and the lemmatization dictionary. We looped on the word dictionary, and linked series of conditions that will be applied according to the type of each word (i.e. lemmatized word, capitalized word, stemmed word…) and its existence in the loaded embeddings: the first condition is that of the existence of a word in the two embeddings, in this case we apply the following formula:

$$V_{Embeddings} = \frac{V_{Embeddings1}(w) + (V_{Embeddings2}(w) + \overline{V}_{Embeddings1} - \overline{V}_{Embeddings2})}{2} \quad (1)$$

$w$ : Index of the word in the corpus
$V_{Embeddings}$ : Vector of concatenated Embeddings
$V_{Embeddings1}$ : First Embedding vector
$V_{Embeddings2}$ : Second Embedding vector
$\overline{V}_{Embeddings1}$ : Mean of the first Embedding vector
$\overline{V}_{Embeddings2}$ : Mean of the second Embedding vector

Once calculated, the new vector is stored in the main embedding matrix with the appropriate word index in the corpus. The following formula summarizes the process of storing the new information to the embedding matrix:

$$M_{Embeddings}(dict_{words}(w)) = V_{Embeddings} \quad (2)$$

$w$ : Index of the word in the corpus
$M_{Embeddings}$ : Matrix of Embeddings
$dict_{words}$ : Dictionnary of words (corpus)
$V_{Embeddings}$ : Vector of concatenated Embeddings

The result of (1) must be a valid value (if it returns a null value, (2) is not applicable), which means that if the word doesn't exist in the both embeddings vectors then, the word is considered as an unknown word and it have to be stored in the unknown vector. Mean values of the embeddings represent major weights that makes the concatenation process balanced, especially for heavy embeddings types (i.e. contains millions in indexes).

The second condition of the algorithm is the existence of the word in the first embedding vector, in this case, only the word vector is stored in the matrix of embeddings. The last condition is for the existence of the word in the second embedding vector, the following formula refers to this condition that is before storing in the embedding matrix:

$$V_{Embeddings} = V_{Embeddings1}(w) + (V_{Embeddings2}(w) + \overline{V}_{Embeddings1} - \overline{V}_{Embeddings2}) \quad (3)$$

$w$ : Index of the word in the corpus
$V_{Embeddings}$ : Vector of concatenated Embeddings
$V_{Embeddings1}$ : First Embedding vector
$V_{Embeddings2}$ : Second Embedding vector
$\overline{V}_{Embeddings1}$ : Mean of the first Embedding vector
$\overline{V}_{Embeddings2}$ : Mean of the second Embedding vector

Finally, for the unknown vector, we multiply the Zero matrix defined in the beginning with -1, so that we can assign a unified index that is specified for unknown words, and then store it in the embedding matrix. And so, we finish filling the embedding matrix while respecting the number of words in the corpus and the size of the embedding defined in the initialization of the algorithm. The next step is to test this embedding concatenation algorithm so that we can evaluate the performance of this mechanism in a neural network context.

For this purpose, we propose four pre-trained embedding models that have as a maximum embedding size of 300:

- glove840B300d: GloVe is an unsupervised learning algorithm for obtaining vector representations for words. Training is performed on aggregated global word-word co-occurrence statistics from a corpus, and the resulting representations showcase interesting linear substructures of the word vector space. This is a Common Crawl version that has 840B tokens, 2.2M vocabularies, cased, 300d vectors.

- GoogleNews-vectors-negative300: A pre-trained vectors trained on part of Google News dataset (about 100 billion words). The model contains 300-dimensional vectors for 3 million words and phrases.

- paragram_300_sl999: A 300 dimensional Paragram embeddings tuned on SimLex999 dataset.
- wiki-news-300d-1M-subword: 1 million-word vectors trained on Wikipedia 2017, UMBC webbase corpus and statmt.org news dataset

Thus, a recurrent model was implemented to test this architecture for a prediction of tourist sentiment towards shopping places: the first layer contains input of the data and it has 60 as maximum length, the second layer is for embeddings and it contains number of words, the embedding matrix, and the size of this matrix, this layer is actually the core of this model which we are pleased to evaluate the model and make comparison with the BERT embedding.

Two regularization layers are initialized (Spatial dropout with one dimension and a regular dropout), and then a Bidirectional Long Short-Term Memory (BiLSTM) [19] layer is implanted in the model with a size of 512 while allowing sequence returning, this layer is stored in a new variable; a Bidirectional Gated Recurrent Unit (BiGRU) [20] layer is created with a size of 256 and sequence returning enabled, and stored in another variable.

These two recurrent architectures are supported by two distinct dropout layers, and these two defined variables are transferred to two distinct GlobalMaxPooling1D layers and then concatenated and connected to the prediction layer with a SoftMax activation function. The model is trained on a range of learning rates that is close to the optimal one found in the BERT section (between $10^{-8}$ to $10^{-2}$), with Adam as an optimization function.

The optimal learning rate isn't performing well on decreasing loss values, and this is due to the optimization function that have to be replaced, and thus, we propose to evaluate a bunch of optimization functions with setting the learning rate value [21] found in the figure above, and switching the concatenated embeddings defined. The proposed optimization functions are: SGD, SGD with momentum of 0.9, AdaGrad, AdaDelta, Adam). The figures below show the performance of each optimizer while switching the 4 embedding sentences.

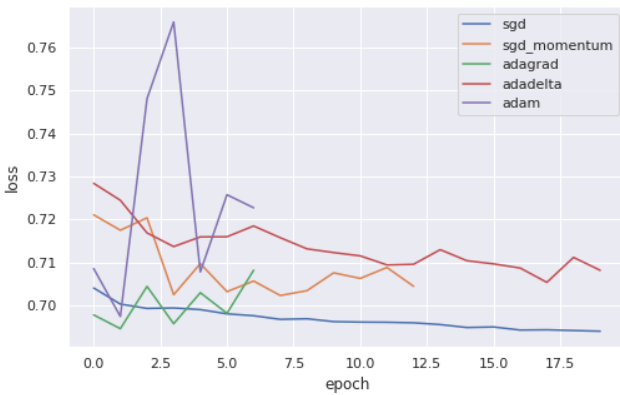

Fig. 3. Optimizers performances for the GloVe and GoogleNews

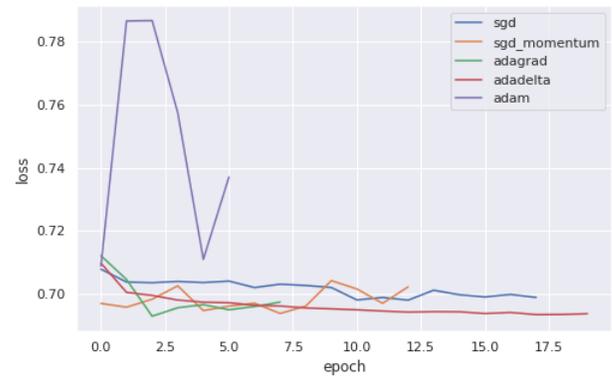

Fig. 4. Optimizers performances for the GloVe and Wiki-News

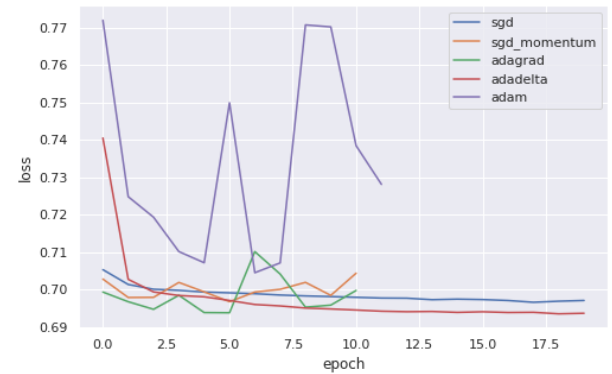

Fig. 5. Optimizers performances for the GloVe and Paragram SL999

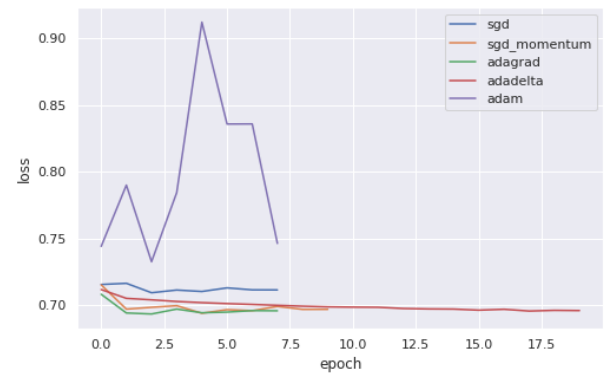

Fig. 6. Optimizers performances for the GoogleNews and Wiki-News

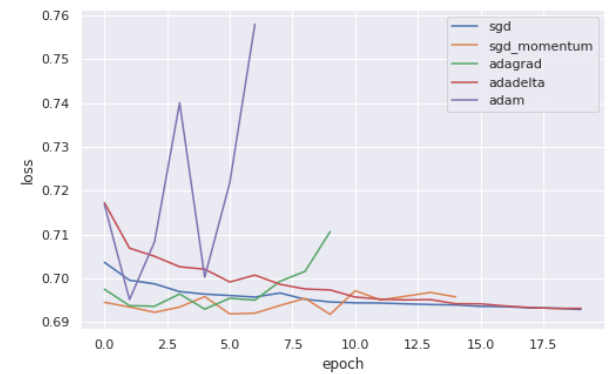

Fig. 7. Optimizers performances for the GoogleNews and Paragram SL999

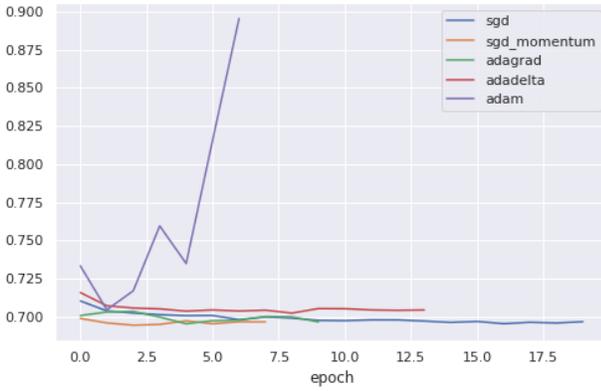

Fig. 8. Optimizers performances for the Wiki-News and Paragram SL999

As can be seen in the plots above, Adam optimizer is performing bad results in all embedding concatenations, the curve has an almost sinusoidal shape, and doesn't succeed to converge, as well as for the Adagrad and SGD with momentum in all the embedding combinations, the convergence stops at early epochs but with less loss values in comparison with Adam performances.

For the SGD curve, the convergence exceeds 20 epochs in all the concatenations, which gives the idea to assign more epochs for the model to learn even more, and decrease loss values, which is applicable for all the cases shown in the graphs above, meanwhile, we are seeking for the best concatenation embedding that can make the model powerful in term of concatenations and learning as well. For the Glove and Google-News concatenation (Fig.3), Adadelta optimizer is reaching the minimum loss value during learning, and it has a tendency to be decreased if we give the model more epochs to learn. The Glove-Wiki plot (Fig.4) gave SGD with momentum as the best optimizer in term of seeking for the minimum loss value, and in term of convergence rapidity despite its stationarity.

Similarly, for the Glove-Paragram combination (Fig.5), which gives almost the same results for the two optimizers mentioned in the first place with superpositions on the two curves, while for the other optimizers, the interpretations remain the same (Adam is always poor and fast convergence and not sufficient for the other optimizers).

In Fig.6, the 4 plots (SGD, SGD Momentum, Adagrad, Adadelta) are performing almost the same results in term of decreasing losses, with little laps in convergence speed, where SGD is always gives tendency to learn even more if the epochs are more than 20.

In another hand, the Google-News Paragram combination in Fig.7 has performed different results than other plots presented above, the Adadelta gave the worst result with a curving allure that is close to the Adam, while the SGD with momentum and Adagrad gave the minimum loss values with stationary curves.

Finally, in Fig.8, we can notice that the Adadelta optimizer curve has the allure of decreasing if we increase epochs values, and it can give excellent results, for the SGD, the curve is almost constant starting from the $1^{st}$ epoch, for the other optimizers the curves are randomly reaching the highest and the lowest loss values without any regular learning rules.

## IV. DISCUSSIONS

In the previous sections, we have presented two comparative approaches that alternate on the tokenization of words and sentence embedding applied on neural network architectures that has as a goal the prediction of tourist's sentiments towards "Jamaa El-Fena" place in Morocco. Therefore, we find that the methodology followed in this work has given hidden insights in term of the performance of combined embedding pre-trained models with the combination of BiLSTM and BiGRU layers in one neural networks architecture, and also the strength of the BERT model with all its components (from word embedding to the model), and the filter we applied when preprocessing the data at the beginning, focusing on the most popular shopping place on TripAdvisor, and which contains a huge amount of information provided by tourists with their ratings that are the basis of this analysis. In one hand, the use of BERT has shortened the steps of the elaboration of the predictive model, since it is based on the encoders which are the core of this architecture, and which gives results relevant with the results of tokenization of BERT which is based on a system which belongs to the latter, and then, the idea of dividing the data at 90% for the training set gave the BERT model to discover more attributes that act on the tourist's sentiments; thus, the technique used for the search of the optimal learning rate allowed us to test the training on several values of the learning rate, focusing on the decrease of the loss values of the model on 20 epochs: the optimal result for this model which is learned on the TripAdvisor data and with an AdamW optimizer on a learning rate of $2*10^{-4}$ gave a score accuracy of 75%. This result can be considered as a compact model [22] which is based on the tools provided by BERT (tokenization, optimizer, ...), and explains the power of this model on several learning topics including the application on the sentiment analysis [23], we can admit that this model is powerful in terms of speed of learning since the elements of this model are very relevant among others [24], and also the coherence between the embedding of BERT, and the BERT model [25]. On the other hand, the use of the concatenation of the embeddings performed unexpected results in particular in the part of the test of the optimizers for each combination of embeddings, we note that the results differ according to each combination, which explains that each pretrained model of embedding has its own properties such as the data it came from [26] and also the formulas we have proposed for the embedding to be well balanced during the concatenation process, in fact the concatenation was done on the basis where the first embedding is chosen as the vector of longer embedding than the second, we started with the Glove which is the longest vector of this set of embedding vectors that we introduced in the previous section, so the use of the average allowed the second embedding vector to be balanced and prepared to be concatenated [27] with the first embedding which is obviously wider than this one. In another context, the use of recurrent networks with two layers of memory to prevent Vanishing Gradient Descent has been a good alternative to overcome all the problems that arise when learning on this kind of data which has a very high semantics with multiclass labels, which makes the prediction task less efficient in terms of accuracy and speed; this alternative gave very efficient results while competing with the BERT model which is based on transformers. Thanks to the idea of launching the model into multiple optimizers while setting one learning rate defined in the testing phase, the plots presented above gave a global vision on the progress of the learning on each combination of embedding where we found that the optimizer Adam is the worst for all the combinations, which is abnormal for an optimizer which is very fast and efficient, the plots presented above gave a global vision on the progress of the learning on each combination of embedding where we found that the optimizer Adam is the worst for all the combinations, which is abnormal for an optimizer which is very fast and efficient; also we note that most of the curves presented are disturbed and do not give interesting results on the data used in this study, with the exception of the Google-News and Wiki-News combination which gave very close results between them and tends that learning can improve if we consider using the SGD model, this explains that the concatenation of two embeddings that contain "News" content gave the model a hold of a giant index embedding from which it adapts with the data which has the tourist aspect and which is informative at the base.

## V. Conclusion

In order to improve the results of sentiment analysis models in a tourism context, the objective of this paper is to compare two approaches to automatic natural language processing from which we have developed an algorithm for the concatenation of sentence embeddings applied to a recurrent neural network composed of BiLSTM and the BiGRU that will prevent vanishing gradient descent during learning; and we compared it with a BERT model. This work will emphasize several topics that will be focused on the deployment of the concatenation of pre-trained models on recent predictive models based on encoders and transformers.